\documentclass[letterpaper, 10 pt, conference]{ieeeconf}  

\IEEEoverridecommandlockouts       
\title{\LARGE \bf
Search at Scale: Improving Numerical Conditioning of Ergodic Coverage Optimization for Multi-Scale Domains
}

\author{Yanis Lahrach$^{1*}$, Christian Hughes$^{2*}$, and Ian Abraham$^{2}$%
\thanks{$^{1}$Yanis Lahrach is with the Department of Applied Mathematics, UCLouvain (Universit\'e catholique de Louvain), Louvain-la-Neuve, Belgium.
{\tt\small yanis.lahrach@student.uclouvain.be}}%
\thanks{$^{2}$Christian Hughes and Ian Abraham are with the Department of Mechanical Engineering,
Yale University, New Haven, CT 06520, USA.
{\tt\small \{christian.hughes, ian.abraham\}@yale.edu}}%
\thanks{$^{*}$Equal contribution.}
}

\usepackage{times}
\usepackage{cite, calc}
\usepackage{multicol}
\usepackage[bookmarks=true]{hyperref}
\usepackage{amsmath,amssymb}

\usepackage{graphicx}
\usepackage{float}
\usepackage{textcomp}
\usepackage{xcolor}
\usepackage{balance}
\usepackage{lipsum}
\usepackage{cuted}
\let\labelindent\relax
\usepackage{enumitem}
\usepackage{algorithm}
\usepackage{algorithmic}

\newtheorem{lemma}{Lemma}
\newtheorem{corollary}{Corollary}
\newtheorem{prop}{Proposition}
\newtheorem{remark}{Remark}
\newtheorem{assumption}{Assumption}
\newtheorem{definition}{Definition}

\usepackage{booktabs}  

\makeatletter
\newcommand{\linebreakand}{%
  \end{@IEEEauthorhalign}
  \hfill\mbox{}\par
  \mbox{}\hfill\begin{@IEEEauthorhalign}
}
\makeatother

\newcommand{\cE}{\mathcal{E}}

\begin{document}

\maketitle
\thispagestyle{empty}
\pagestyle{empty}

\begin{abstract}

    Recent methods in ergodic coverage planning have shown promise as tools that can adapt to a wide range of geometric coverage problems with general constraints, but are highly sensitive to the numerical scaling of the problem space.    
    The underlying challenge is that the optimization formulation becomes brittle and numerically unstable with changing scales, especially under potentially nonlinear constraints that impose dynamic restrictions, due to the kernel-based formulation.
    This paper proposes to address this problem via the development of a scale-agnostic and adaptive ergodic coverage optimization method based on the maximum mean discrepancy metric (MMD). 
    Our approach allows the optimizer to solve for the scale of differential constraints while annealing the hyperparameters to best suit the problem domain and ensure physical consistency.  
    We also derive a variation of the ergodic metric in the log space, providing additional numerical conditioning without loss of performance. 
    We compare our approach with existing coverage planning methods and demonstrate the utility of our approach on a wide range of coverage problems.
            
\end{abstract}

\section{INTRODUCTION}

    Effective coverage planning is at the core of many robotic automation problems such as mapping, exploration, and inspection. 
    However, current methods require hyper-specific tuning to the coverage problem, resulting in a wide range of algorithmic adaptations. 
    For example, coverage planning at the micro-scale level, e.g., in part inspection, requires highly precise control and planning \cite{wahab2014}, while macro-level coverage planning, e.g., for weather monitoring, requires localized measurements over large spaces while satisfying differential velocity constraints \cite{luca2005}.  

    Ergodic trajectory optimization has emerged as a popular general approach for coverage planning that minimizes the distance between the time-averaged statistics of a trajectory and a target spatial distribution while integrating differential dynamics constraints \cite{abraham2020}. 
    More recent advancements have expanded the use-case of ergodic coverage to more complex spaces via kernel-based methods \cite{sun2024, sun2025, hughes2025}. 
    However, due to their reliance on kernel functions, these approaches are heavily dependent on the appropriate selection of hyperparameters and domain scale, leading to brittleness and numerical conditioning issues at optimization time. 
    Furthermore, tuning the hyperparameters also alters the physical interpretation of coverage, i.e., a robot's sensor footprint, which requires careful consideration as inappropriate selection can lead to unstable optimizations and trajectories that do not prioritize coverage sufficiently. 

\begin{figure}[t!]
\centering
\includegraphics[width=\linewidth]{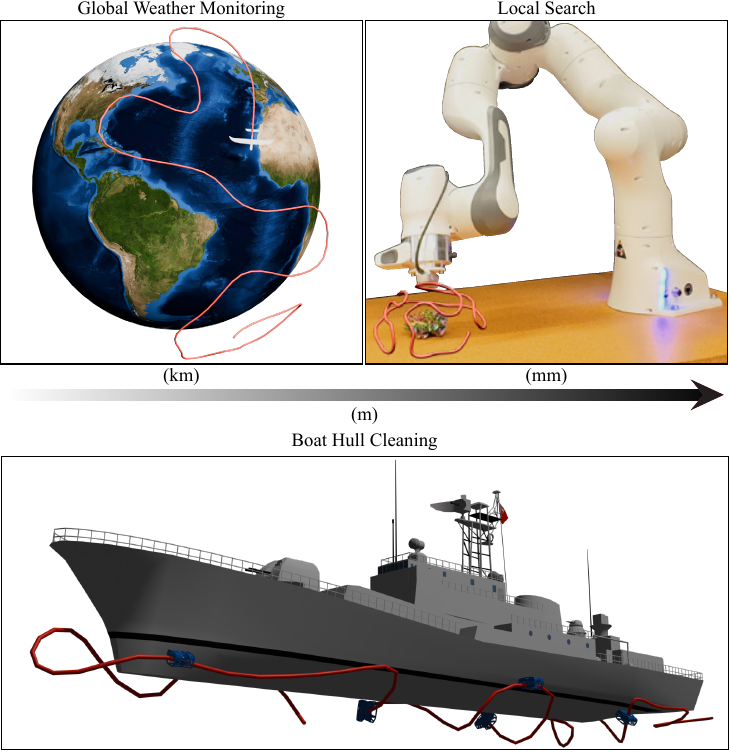}
\vspace{-15pt}
\caption{\textbf{Scale-Agnostic Ergodic Search.} Our single, scale-invariant optimizer enables proportional coverage guarantees for tasks spanning vastly different spatial scales. A trajectory is generated for a Franka Emika Panda robot inspecting a small rock (right) and for a drone surveying the Atlantic Ocean for global-scale weather monitoring (left).}
\label{fig:abstract}
\end{figure} 

    In this paper, we present a reformulation of the ergodic coverage trajectory optimization problem that is invariant to the spatial scale of the search space, decoupling hyperparameters from the physical scale of the task, without compromising performance.
    Our approach uses the maximum-mean discrepancy (MMD) metric to compute ergodicity (derived in \cite{hughes2025}) with various modifications to ensure smooth and scalable optimization. 
    First, we propose an automatic normalization of the search domain that is decoupled from the dynamic constraints. 
    Second, we allow the solver to adapt the time between differential constraints, e.g., between trajectory points, to best suit the coverage problem domain, enabling improved coverage in spatially sparse domains. 
    Third, we propose annealing techniques for automatic hyperparameter selection to adapt kernel length-scales to the given footprint of a robot without loss of physical meaning.
    Last, we propose a log variation of the MMD metric that is robust to the problem scale. 
    This new approach to ergodic MMD optimization enables practical ergodic coverage for applications at any scale, from silicon coating in microelectronics to oceanic weather monitoring (as shown in Fig.~\ref{fig:abstract}).

    \emph{In summary, the primary contribution of this work is an ergodic trajectory optimizer that}:
    \begin{itemize}
        \item improves numerical conditioning in coverage planning across a range of geometries and spatial scales, 
        \item maintains strict adherence to continuous dynamical constraints, and
        \item facilitates reliable hyperparameter selection that generalizes across diverse scales.
    \end{itemize}

    The paper is outlined as follows: Section~\ref{sec:related_work} reviews related work, Section~\ref{sec:prelims} covers preliminary background information on kernel-based methods and ergodic coverage planning, Section~\ref{sec:emmd} derives the modified ergodic coverage optimization, and Sections~\ref{sec:results} and~\ref{sec:conclusion} contain the results and conclusions, respectively.

\section{RELATED WORK} \label{sec:related_work}

\subsection{Robotic Exploration}

Approaches to autonomous exploration generally prioritize one of two main objectives: coverage over the search space and maximal information gain. 
Coverage-based methods aim to generate trajectories that ensure every reachable region in the search environment is visited. 
Traditional approaches often solve this as a Traveling Salesperson Problem (TSP) over a discretized grid, where visiting a cell signifies full coverage of that region \cite{laporte1987, applegate2006}.
While these methods offer theoretical guarantees of complete coverage, they often lack compatibility with dynamic and kinematic constraints used in robotic systems.

Information-driven optimization approaches have emerged as a popular strategy in robotics due to their prioritization of efficient exploration over the search space \cite{bourgault2002}.
These approaches optimize the robot's trajectories according to a spatial information distribution that represents the predicted information gain of visiting different regions \cite{amigoni2010}.
Early work in this field involved integrating probabilistic models into control policies to minimize uncertainty across different regions of the search space \cite{thrun1997}.
However, while successful for short-term tasks with reliable information, these approaches can struggle to provide coverage of the search space over longer time-horizons and in uncertain scenarios.
Recent work in ergodicity-based exploration methods have been shown to compensate for this lack of coverage in information-theoretic settings. 




\subsection{Ergodic Search} 

Ergodic coverage methods seek to minimize the distance between a robot's time-averaged trajectory and a given spatial information distribution. 
Ergodicity-based methods have demonstrated the ability to produce trajectories with a theoretical guarantee of complete coverage over continuous domains as time approaches infinity \cite{abraham2020}.
Traditional ergodic optimizers minimize a spectral ergodic metric based on Fourier basis functions that has been shown to generate effective coverage trajectories for a wide variety of information-gathering tasks \cite{mavrommati2018, sartoretti2021, lee2024}.
However, these standard approaches to ergodicity-based planning are limited to bounded, well-defined Euclidean spaces, which limits their applicability in environments with complex geometries. 

Recent advancements have enabled ergodic coverage in environments with complex geometries using a variety of approaches. 
These methods either derive basis functions from the domain \cite{sun2024fast}, solve partial differential equations (PDEs) for ergodic control policies \cite{ivic2022}, or numerically approximate basis functions via the Laplace-Beltrami Operator \cite{bilaloglu2024tactile}. 
However, each of these solutions requires specific tuning to the domain: \cite{sun2024fast} does not apply to unstructured domains (meshes), \cite{ivic2022} requires computing a PDE to obtain a local ergodic controller and does not enable trajectory planning, and \cite{bilaloglu2024tactile, belkin2008} requires approximation of basis functions that can become computationally challenging. 


More recent work has found success in generating ergodic coverage trajectories over arbitrarily-shaped domains by leveraging ideas from optimal transport~\cite{sun2025} and two-sample maximum mean discrepancy (MMD)~\cite{hughes2025}. 
These methods formulate ergodic coverage as a distribution matching problem between a general form of the time-averaged statistics and domain samples rather than any structured domain definition. 
However, these approaches are highly sensitive to proper selection of hyperparameters (see Fig.\ref{fig:kernel_comp}) as well as domain scales (see Fig.\ref{fig:compare_3d}) that lead to significant performance degradation. 
In this paper, we propose a series of modifications to the ergodic MMD metric that decouple the dual responsibilities of kernel hyperparameters and improve the condition of the problem at arbitrary spatial scales. 




\section{PRELIMINARIES}\label{sec:prelims}

In this section, we recall (i) time–averaged visitation statistics and ergodicity, (ii) kernel mean embeddings, and (iii) a Maximum Mean Discrepancy (MMD) objective for coverage.

\subsection{Setup and notation}

Let $\Omega \subseteq \mathbb{R}^d$ denote a measurable search domain with Borel $\sigma$–algebra $\mathcal{B}$, and let $\mathcal{P}(\Omega)$ be the set of Borel probability measures on $\Omega$.
A robot evolves on a state space $\mathcal{X}\subseteq\mathbb{R}^n$ with 
a differentiable map $g:\mathcal{X}\to\Omega$ (e.g., task–space pose or surface parameterization) and returns the visited location at time $t$:
\[
\omega^x_t \;:=\; g(x_t)\in\Omega.
\]
Over a finite horizon $T\in\mathbb{N}$, we write the discrete trajectory as $x=\{x_0,\dots,x_{T-1}\}$ \footnote{One can view the trajectory as an ordered list of sensor locations.} that are constrained by the function 
\begin{equation} \label{eq:physics_constr}
    f(x_{t+1}, x_t) = 0
\end{equation}
where $f$ is a function that enforces physical consistency, e.g., dynamics and initial/final conditions.

\subsection{Time–averaged visitation and ergodicity}\label{subsec:ergodicity}

\begin{definition}[Time–averaged (empirical) visitation]
The visitation statistics of a trajectory $x$ in $\Omega$ are
\begin{equation}
  \rho_{x,T}
  \;=\;
  \frac{1}{T}\sum_{t=0}^{T-1}\delta_{\omega^x_t}
  \;\in\;\mathcal{P}(\Omega),
  \qquad
  \omega^x_t = g(x_t),
  \label{eq:time_avg_stats}
\end{equation}
where $\delta_{\omega}$ denotes the Dirac measure at $\omega$.
\end{definition}

\begin{definition}[Ergodicity]
A (possibly infinite) trajectory $x$ is \emph{ergodic} with respect to a target utility $\mu\in\mathcal{P}(\Omega)$ if, for every continuous test function $\phi\in\mathcal{C}(\Omega)$,
\[
\lim_{T\to\infty}\int_\Omega \phi(\omega)\,d\rho_{x,T}(\omega)
\;=\;
\int_\Omega \phi(\omega)\,d\mu(\omega).
\]
Equivalently: time averages equal $\mu$–weighted space averages.
\end{definition}

In finite time, we aim to design trajectories whose empirical statistics $\rho_{x,T}$ match a given $\mu$ as closely as possible while satisfying physical constraints.
To do so, an ergodic metric $\mathcal{E} : \mathcal{X}^T \times P(\Omega) \to \mathbb{R}$ is optimized for trajectories $x$ subject to physical constraints~\eqref{eq:physics_constr}. More specifically, the ergodic coverage problem is given as 
\begin{equation} \label{eq:erg_opt}
    \min_{x} \mathcal{E}(x ; \mu) \quad\text{s.t.  } \forall t \, f(x_{t+1},x_t) = 0, h(x) \le 0
\end{equation}
where $f$ enforces physical constraints and $h$ are inequality constraints. 
In the following section, we derive a variation of the ergodic metric based on maximum mean discrepancy and kernel mean embeddings that enables us to specify coverage problems on a wide variety of geometries and domains. 

\subsection{Kernel mean embeddings}\label{subsec:kme}

    We first define kernel mean embeddings of a distribution. 
    A function $k:\Omega\times\Omega\to\mathbb{R}$ is (symmetric) positive–definite if, for any $m\in\mathbb{N}$, points $\{\omega_i\}_{i=1}^m\subset\Omega$ and coefficients $\{\alpha_i\}_{i=1}^m\subset\mathbb{R}$,
    $\sum_{i,j=1}^m \alpha_i\alpha_j\,k(\omega_i,\omega_j)\ge 0$.
    To such a kernel is associated a unique reproducing kernel Hilbert space $(\mathcal{H}_k,\langle\cdot,\cdot\rangle_{\mathcal{H}_k})$ with feature map $\phi:\Omega\to\mathcal{H}_k$, $\phi(\omega)=k(\cdot,\omega)$, satisfying the reproducing property $f(\omega)=\langle f,\phi(\omega)\rangle_{\mathcal{H}_k}$ for all $f\in\mathcal{H}_k$.

\begin{assumption}[Stationary, isotropic kernels]
We use stationary, isotropic kernels of the form
\begin{equation}
k(\omega,\omega')=\kappa\!\left(\frac{-d(\omega,\omega')}{h}\right),\qquad h>0,
\label{eq:stationary_kernel}
\end{equation}
with length (bandwidth) parameter $h$ and distance metric $d : \Omega \times \Omega \to \mathbb{R}$. A canonical choice is the Gaussian/RBF kernel $k(\omega,\omega')=\exp\!\big(-\|\omega-\omega'\|_2^2/h\big)$.
\end{assumption}

\begin{definition}[Kernel mean embedding \cite{gretton2012}]
For $p\in\mathcal{P}(\Omega)$, the kernel mean embedding is the Bochner integral
\begin{equation}
m_p(\cdot)\;:=\;\mathbb{E}_{\omega\sim p}\big[k(\cdot,\omega)\big]\;\in\;\mathcal{H}_k,
\label{eq:kme_def}
\end{equation}
which represents $p$ as a smooth “coverage field’’ in $\mathcal{H}_k$.
\end{definition}

\begin{remark}[Preview: scale sensitivity]
Because \eqref{eq:stationary_kernel} depends on $\|\omega-\omega'\|/h$, a uniform change of physical units alters the effective resolution unless $h$ is co–tuned. Section~\ref{sec:emmd} introduces a normalization that decouples similarity from units and lets us reuse a \emph{single numerical} $h$ across scales.
\end{remark}

\subsection{Maximum Mean Discrepancy (MMD)}\label{subsec:emmd_prelim}

\begin{definition}[Maximum Mean Discrepancy]
Given $p,q\in\mathcal{P}(\Omega)$ with embeddings $m_p,m_q\in\mathcal{H}_k$, the squared MMD is
\begin{equation}
\mathrm{MMD}_k^2(p,q)\;:=\;\big\|m_p-m_q\big\|_{\mathcal{H}_k}^2.
\label{eq:mmd_rkhs}
\end{equation}
\end{definition}

\begin{prop}[Expectation form of MMD \cite{hughes2025}]
Let $\mathbb{E}_{p,q}[k]:=\iint k(\omega,\omega')\,dp(\omega)\,dq(\omega')$. Then
\begin{equation}
\mathrm{MMD}_k^2(p,q)
= \mathbb{E}_{p,p}[k]\;-\;2\,\mathbb{E}_{p,q}[k]\;+\;\mathbb{E}_{q,q}[k].
\label{eq:mmd_expectation}
\end{equation}
If $k$ is characteristic, then $\mathrm{MMD}_k(p,q)=0\iff p=q$.
\end{prop}

\begin{prop}[Ergodic MMD objective]
Let $\rho_{x,T}$ be the time–averaged visitation measure from \eqref{eq:time_avg_stats} and let $\mu\in\mathcal{P}(\Omega)$ be a target utility. The \emph{ergodic MMD} objective
\begin{equation}
\mathcal{E}_k(x;\mu)\;:=\;\mathrm{MMD}_k^2\!\big(\mu,\rho_{x,T}\big)
\label{eq:emmd_objective}
\end{equation}
is minimized when the finite–time statistics of $x$ match $\mu$ in the Hibert space distance.
\end{prop}


\begin{prop}[Finite-sample MMD estimator]
Suppose $\mu$ is represented by samples $\{\omega_i\}_{i=1}^{M}\sim \mu$ and write $\omega^x_t=g(x_t)$. An estimator of \eqref{eq:emmd_objective} is
\begin{equation}
\begin{aligned}
\mathcal{E}_k(x;\mu)
&= \frac{1}{T^2}\sum_{t,t'=0}^{T-1} k(\omega^x_t,\omega^x_{t'})
   - \frac{2}{TM}\sum_{t=0}^{T-1}\sum_{i=1}^{M}\,k(\omega^x_t,\omega_i) \\
&\quad + \frac{1}{M^2}\sum_{i=1}^{M}\sum_{j=1}^{M}\,k(\omega_i,\omega_j),
\end{aligned}
\label{eq:mmd_finite}
\end{equation}
where the last term is constant w.r.t. $x$ and can be precomputed.
\end{prop}

The finite–sample objective in \eqref{eq:mmd_finite} has three parts:
(i) the self–similarity term $\frac{1}{T^2}\!\sum_{t,t'} k(\omega^x_t,\omega^x_{t'})$
penalizes trajectories that revisit the \emph{same} places (it encourages spatial
\emph{dispersion});
(ii) the cross term $-\frac{2}{TM}\!\sum_t\sum_i \,k(\omega^x_t,\omega_i)$
pulls the trajectory toward high–utility samples (it drives \emph{coverage});
(iii) the last term depends only on $\mu$ and is constant w.r.t.\ $x$.
The kernel bandwidth $h$ sets the robot's effective visitation footprint:
large $h$ rewards broad, coarse coverage; small $h$ rewards fine, local coverage.

We optimize \eqref{eq:mmd_finite} subject to dynamics and path constraints as described in~\eqref{eq:erg_opt}.
For clarity we first write the gradient entirely in the search space $\Omega$.
If $k\in C^1$, the map $x\mapsto \cE_k(x;\mu)$ is differentiable via the chain rule. For the RBF kernel $k(\omega,\omega')=\exp(-\|\omega-\omega'\|^2/(2h))$, the derivative of $\mathcal{E}_k$ with respect to $x_t$ is given as 
\begin{equation}
\frac{\partial \cE_k}{\partial x_t}
= \frac{\partial g}{\partial x_t}^\top\,\frac{\partial \cE_k}{\partial \omega^x_t}.
\end{equation}
Expanding out the latter term gives the following
\begin{equation}
\begin{aligned}
\label{derivative}
\frac{\partial \cE_k}{\partial \omega^x_t}
&= \frac{2}{T^2}\sum_{t'} 
   \frac{\omega^x_t-\omega^x_{t'}}{h}\,
   k(\omega^x_t,\omega^x_{t'}) \\[-1pt]
&\quad - \frac{2}{TM}\sum_{i}
   \frac{\omega^x_t-\omega_i}{h}\,
   k(\omega^x_t,\omega_i).
\end{aligned}
\end{equation}
Because the kernel is written in physical distances, a unit scaling $\omega \mapsto e\,\omega$ changes the factors in \eqref{derivative}: the displacement $(\omega_s-\omega^2_t)$ scales linearly by $e$, while $k(eu,ev)=\exp(-e^{2}\|u-v\|^{2}/h)$ changes exponentially.
Consequently the gradient magnitude behaves like $\sim (e/h)\,\exp(-e^{2}\cdot/h)$: for large $e$ (fixed $h$) it decays toward zero, and for small $e$ it grows—thus the optimization varies with scale.

\section{ERGODIC COVERAGE AT SCALE}\label{sec:emmd}

We improve the numerical conditioning of kernel-based ergodic optimization without altering the physical meaning of constraints. The key idea is to (i) evaluate kernel similarity in a \emph{dimensionless} domain, (ii) adapt differential dynamics and path constraints via optimized time, and (iii) derive a log-sum-exp (LSE) surrogate that yields smooth, size-robust gradients. A practical mapping from a desired \emph{physical} footprint to a normalized bandwidth is given later in Sec.~\ref{subsec:annealing_footprint}.

\subsection{Scale Normalization and Physical Consistency}\label{subsec:scale}

Let $\Omega\subset\mathbb{R}^d$ be the search domain and $\{\omega_i\}_{i=1}^M\subset\Omega$ target samples with nonnegative weights $\{\pi_i\}_{i=1}^M$ (normalized so $\sum_i\pi_i=1$). We define the environment extent
\[
e \;=\; \max_{\ell\in\{1,\dots,d\}}
\Big(\max_i[\omega_i]_\ell-\min_i[\omega_i]_\ell\Big)\;>\;0,
\]
and normalized coordinates $\hat\omega=\omega/e\in\hat\Omega$. For a trajectory $x=\{x_t\}_{t=0}^{T-1}$ in physical units, we have $\hat g(x_t)=\hat{ \omega}^x_t= 
\omega_t^x / e$.

We compute kernel similarity on $\hat\Omega$ with a bandwidth $h>0$,
\begin{equation}\label{eq:scaled-kernel}
k_h(\hat u,\hat v)=\exp\!\big(-\|\hat u-\hat v\|_2^2/h\big).
\end{equation}
In physical coordinates, this is $k_h(u,v)=\exp(-\|u-v\|_2^2/(h\,e^2))$. Because distances in $\hat\Omega$ are $\mathcal{O}(1)$, the \emph{same numerical} $h$ can be re-used across very different physical scales. When a \emph{fixed physical footprint} is required, we map it to a normalized bandwidth and anneal toward the physically consistent footprint.

\subsection{Bandwidth Annealing and Physical Footprint}
\label{subsec:annealing_footprint}

Keeping the \emph{normalized} bandwidth $h$ fixed as the environment scale $e$ changes is equivalent to changing a \emph{physical} bandwidth to
\[
h_{\rm phys}=h\,e^2,
\]
so the effective physical footprint of visitation grows/shrinks with $e$. In other words, a constant $h$ on the normalized domain does \emph{not} preserve the robot's physical sensing/actuation footprint.

To preserve physical meaning, we choose a target physical bandwidth $h_{\rm phys}^\star$ (fixed across scales) and set
\[
h_{\rm norm}^\star \;=\; \frac{h_{\rm phys}^\star}{e^2}.
\]

If we start optimization directly with a small $h_{\rm norm}^\star$ on a large domain, then $k_h(u,v)\approx 0$ for most pairs and the terms are dominated by self-similarity; gradients are nearly zero and the solver stalls.  
We therefore use a \emph{continuation} (annealing) by beginning with a large normalized bandwidth $h_0$ that produces wide, informative kernels, and decrease it geometrically to $h_{\rm norm}^\star$.


\paragraph*{Annealing schedule}
In practice, we define an annealing schedule that the optimization follows to facilitate continual solver cost reduction. Let $e$ be the environment extent, $h_{\rm norm}^\star = h_{\rm phys}^\star/e^2$ the target normalized bandwidth, and choose an large initial $h_0$ start, e.g., \( h_0=0.05\).
Over $K$ outer iterations, we use a geometric series
\begin{equation} \label{eq:annealing}
h_k
\;=\;
h_0\left(\frac{h_{\rm norm}^\star}{h_0}\right)^{\!\frac{k}{K-1}},
\qquad k=0,\dots,K-1,    
\end{equation}
optimizing the same loss at each $h_k$ and warm-starting from the previous solution. 
The overall result is that the solver makes meaningful updates to the solution and places trajectory visitation points in key regions of interest while eventually converging into a length scale that is physically consistent to the robot. 
Next, we propose a variation to the differential dynamic constraints to adapt according to the solver. 



\subsection{Adaptive Time Differential Constraints}
    To ensure that differential constraints have the ability to scale to the domain size we present a time-adaptive variation of the optimization~\eqref{eq:erg_opt}. 
    Specifically, we add in a time-scaling component $\Delta_t$ to the physical constraints $f$ that governs the behavior between trajectory points $x_{t+1}, x_t$.
    This allows us to write the differential physics constraint as
    \begin{equation}\label{eq:diff_constr}
        f(x_{t+1}, x_t, \Delta_t) = 0
    \end{equation}
    where $\Delta_t$ can be optimized. 
    For example, given the state vector $x_t=[p_t^\top, v_t^\top]$ where $p_t$ is the position and $v_t$ is a bounded velocity, the constraint~\eqref{eq:diff_constr} is written with Euler integration as 
    $$
        f(x_{t+1}, x_t, \Delta_t)=p_{t+1} - (p_t + \Delta_t v_t) = 0.
    $$
    Since $\Delta_t>0$, we can transform the set of $\Delta_t$ as a trajectory in the log-space $\log(\Delta) = \{ \log \Delta_t \}_{t=0}^{T-1}$ leading to the modified optimization 
    \begin{equation} \label{eq:logdt_erg_opt}
        \min_{x, \log(\Delta)} \mathcal{E}(x ; \mu) \quad\text{s.t.  } \forall t \, f(x_{t+1},x_t, \Delta_t) = 0, h(x) \le 0.
    \end{equation}
    Optimizing over $\log \Delta_t$ has the added advantage of being numerically smooth without the need for additional constraints and keeping differential physics constraints consistent. 



In practice, an adaptive $\Delta_t$ has several additional advantages that expands the use ergodic coverage planning more challenging problems. For instance, (i) \emph{Resolution where it matters}: non-uniform \(\Delta_t\) increases waypoint density in high-utility/high-curvature zones and reduces aliasing of small structures. 
(ii) \emph{Constraint-compatible}: trajectory velocity bounds remain enforced via \(\|x_{k+1}-x_k\| \le v_{\max}\Delta_t\).\footnote{More complex integration schemes can be used without loss of generality.}
(iii) \emph{Scale-robust}: combined with the normalized domain, the spatial sampling adapts to spatial sparsity of regions of exploration across orders of magnitude without retuning.

Our final modification to the ergodic coverage optimization is in the ergodic MMD objective itself which alleviates gradient scaling under optimization. 

\subsection{The Log-Surrogate Ergodic MMD}\label{subsec:logmmd}

    In this section, we define a log variation of the MMD objective that provides a smooth surrogate function. 
    \begin{prop}[Log-surrogate EMMD]\label{prop:logsur}
        Let $\hat \Omega$ be the problem domain with $\mu \in \mathcal{P}(\hat \Omega)$ and $\rho_{x,T} \in \mathcal{P}(\hat \Omega)$ be the target distribution and time-average trajectory statistics respectively. 
        Then, given a locally optimal trajectory $x^\star$, the following holds
        \begin{equation}
        \begin{split}
        \label{eq:log-surrogate}
        \tilde{\mathcal{E}}_h(x;\mu)
        &=\log\!\sum_{t,t'} k_h(\hat\omega_t^x,\hat\omega_{t'}^x)
        -2\log\!\sum_{t,i}\,k_h(\hat\omega_t^x,\hat\omega_i) \\ 
        &\le \mathcal{E}_h(x; \mu).
        \end{split}
        \end{equation}
    \end{prop}
    \begin{proof}
        Taking the log of the ergodic MMD objective in~\eqref{eq:mmd_finite} and using concavity rules for log we get 
        \begin{equation*}
        \begin{split}
            \log \left( \mathcal{E}_h(x; \mu)\right) &\le \mathcal{E}_h(x; \mu) \\ 
                & \le \log\!\sum_{t,t'} k_h(\hat\omega_t^x,\hat\omega_{t'}^x)
            -2\log\!\sum_{t,i}\,k_h(\hat\omega_t^x,\hat\omega_i) \\ 
            &= \tilde{\mathcal{E}}_h(x; \mu)
        \end{split}
        \end{equation*}
        where we remove the third log term as it is a constant and the normalizing factors of $T, M$. 
    \end{proof}

    The surrogate MMD objective has a few properties that makes it advantageous over the MMD objectives. 
    First, the expresssion~\eqref{eq:log-surrogate} is numerically better behaved as it is equivalent to the \emph{log-sum-exp} (LSE) given an RBF kernel. 
    As a result (via logs of sums), produces globally normalized ``soft-attention'' weights, and yields bounded, size-robust gradients on the normalized domain \(\hat\Omega\). 
    Additionally, the log-sum-exp computation prevents over and underflows for especially large numbers that further improves the conditioning of ergodic trajectory optimization.

\begin{lemma}[Nonnegativity and shared global minima]]\label{lem:nonneg}
Let \(m_{\rho},m_{\mu}\in\mathcal{H}_k\) be the kernel mean embeddings of \(\rho_{x,T}\) and \(\mu\).
Then
\[
A=\|m_{\rho}\|_{\mathcal{H}_k}^2,\quad
B=\langle m_{\rho},m_{\mu}\rangle_{\mathcal{H}_k},\quad
C=\|m_{\mu}\|_{\mathcal{H}_k}^2,
\]
and by Cauchy–Schwarz \(B^2\le AC\), hence \(\tilde{\mathcal{E}}_h(x;\mu)=\log(AC/B^2)\ge 0\).
Moreover, \(\tilde{\mathcal{E}}_h(x;\mu)=0\) iff \(m_{\rho}\) and \(m_{\mu}\) are collinear; for \emph{probability} measures and a characteristic kernel, this holds iff \(\rho_{x,T}=\mu\). Therefore \(\tilde{\mathcal{E}}_h\) and \(\mathrm{MMD}_k^2\) share the same global minimum (value \(0\)) and the same set of global minimizers.
\end{lemma}
\begin{proof}
The identities for \(A,B,C\) follow from standard algebra. Cauchy–Schwarz gives
\(B^2\le \|m_{\rho}\|^2\|m_{\mu}\|^2=AC\), hence \(\tilde{\mathcal{E}}_h=\log(AC/B^2)\ge 0\).
Equality in C–S holds iff \(m_{\rho}=\lambda m_{\mu}\) for some \(\lambda\in\mathbb{R}\).
Since the embedding is linear in the measure and both \(\rho_{x,T}\) and \(\mu\) have unit mass, \(\lambda=1\) and \(m_{\rho}=m_{\mu}\) iff \(\rho_{x,T}=\mu\) for a characteristic kernel. 
\end{proof}

\begin{remark}
While \(\tilde{\mathcal{E}}_h\) and \(\mathrm{MMD}_k^2\) have the same global minimum and zero set, they are not generally order-equivalent away from the minimum (logs act termwise). We use \(\tilde{\mathcal{E}}_h\) as a numerically well-conditioned \emph{surrogate} with identical optimal solutions when perfect matching is feasible.
\end{remark}

Interestingly, the log-surrogate MMD objective acts like a soft-attention gradient representation.
Let us define the positive sums
\begin{align*}    
F_{xx}:=\sum_{r,s} k_h(\hat\omega_r^x,\hat\omega_s^x),\qquad
F_{x\mu}:=\sum_{r,i}k_h(\hat\omega_r^x,\hat\omega_i), \\ 
F_{\mu\mu}:=\sum_{i,j}k_h(\hat\omega_i,\hat\omega_j) \hspace{6em}
\end{align*}
and the normalized weights
\[
\alpha_{ts}:=\frac{k_h(\hat\omega_t^x,\hat\omega_s^x)}{F_{xx}},
\qquad
\beta_{ti}:=\frac{\,k_h(\hat\omega_t^x,\hat\omega_i)}{F_{x\mu}}.
\]
Note that \(\alpha_{ts},\beta_{ti}\in(0,1)\) and \(\sum_{s}\alpha_{ts}\le 1\), \(\sum_{t}\alpha_{ts}\le 1\), \(\sum_i\beta_{ti}\le 1\) by global normalization.

\begin{figure}[t!]
    \centering
    \vspace{5pt} 
    \includegraphics[width=\linewidth]{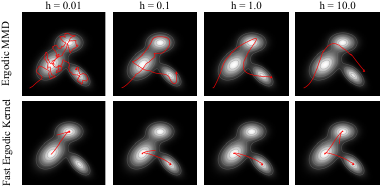}
    \vspace{-15pt}
    \caption{\textbf{Sensitivity of Kernel-Based Methods to Length Scale Selection.} Trajectories are generated for identical 2D environments at spatial scale = 1 with varying length scales (h). Length scale selection can lead to underfitting (if too high) or overfitting (if too low) to the utility samples, resulting in degraded performance or complete optimization failure.}
    \label{fig:kernel_comp}
\end{figure}

\begin{lemma}[Gradient for general metric]\label{lem:grad-general}
Assume \(d\) is differentiable and write \(\nabla_{\hat u} d(\hat u,\hat v)\) for the gradient in the first argument. Then
\begin{equation} \label{eq:grad-general}
\begin{split}
\partial_{\hat\omega_s^x}\,\tilde{\mathcal{E}}_h
= -\frac{1}{h}\sum_{t}\!\big(\alpha_{st}+\alpha_{ts}\big)\,\nabla_{\hat u} d(\hat\omega_s^x,\hat\omega_t^x) \\ 
+\frac{2}{h}\sum_{i}\beta_{si}\,\nabla_{\hat u} d(\hat\omega_s^x,\hat\omega_i).
\end{split}
\end{equation}
Gradients w.r.t.\ the decision variables follow by the chain rule,
\(\partial_{x_s}\tilde{\mathcal{E}}_h
=\big(\tfrac{\partial\hat g}{\partial x}(x_s)\big)^\top
\partial_{\hat\omega_s^x}\tilde{\mathcal{E}}_h\).
\end{lemma}
\begin{proof}
Since \(k_h(\hat u,\hat v)=\exp(-d(\hat u,\hat v)/h)\),
\(\nabla_{\hat u} k_h(\hat u,\hat v)=-(1/h)\,\nabla_{\hat u}d(\hat u,\hat v)\,k_h(\hat u,\hat v)\).
Differentiating \(F_{xx}\) w.r.t.\ \(\hat\omega_s^x\) and accounting for occurrences as first and second argument while dividing by \(F_{xx}\) to obtain \(\partial\log F_{xx}\). Similarly \(\partial\log F_{x\mu}\) follows by the chain rule. Combining all  terms yields \(\partial\tilde{\mathcal{E}}_h=\partial\log F_{xx}-2\,\partial\log F_{x\mu}\). 
\end{proof}

\begin{corollary}[Euclidean specialization]\label{cor:euclid}
If \(d(\hat u,\hat v)=\|\hat u-\hat v\|_2^2\), then \(\nabla_{\hat u} d(\hat u,\hat v)=2(\hat u-\hat v)\) and
\begin{equation}
\label{eq:grad-euclid}
\partial_{\hat\omega_s^x}\,\tilde{\mathcal{E}}_h
= -\frac{2}{h}\sum_{t}\!\big(\alpha_{st}+\alpha_{ts}\big)\,(\hat\omega_s^x-\hat\omega_t^x)
+\frac{4}{h}\sum_{i}\beta_{si}\,(\hat\omega_s^x-\hat\omega_i).
\end{equation}
\end{corollary}

We can show that with the log-surrogate MMD function, we can obtain bounded gradients that are independent of the number of trajectory knot points $T$ and the number domain samples $M$. 

\begin{lemma}[Bounded, size-robust gradients]\label{lem:bounds}
On a bounded normalized domain \(\hat\Omega\) with diameter \(D=\mathrm{diam}(\hat\Omega)\),
\[
\big\|\partial_{\hat\omega_s^x}\log F_{x\mu}\big\|
\le \frac{2D}{h},\qquad
\big\|\partial_{\hat\omega_s^x}\log F_{xx}\big\|
\le \frac{4D}{h},
\]
hence \(\|\partial_{\hat\omega_s^x}\tilde{\mathcal{E}}_h\|\le \frac{6D}{h}\).
These bounds are independent of the number of knots \(T\) and samples \(M\).
\end{lemma}
\begin{proof}
Use \eqref{eq:grad-general} (or \eqref{eq:grad-euclid}) together with
\(\sum_t(\alpha_{st}+\alpha_{ts})\le 2\), \(\sum_i \beta_{si}\le 1\),
and \(\|\hat\omega-\hat\omega'\|\le D\).
\end{proof}

Last, we show that the log-surrogate ergodic MMD objective is scale-invariant. 
\begin{prop}[Scale invariance on \(\hat\Omega\)]\label{prop:scaleinv}
Let \(\omega=e\,\hat\omega\) be a uniform rescaling of the physical domain and \(\hat g(x)=g(x)/e\).
Then \(\tilde{\mathcal{E}}_h\) and its gradients are invariant under rescaling.
\end{prop}
\begin{proof} 
 On $\hat\Omega$, pairwise distances in $k_h$ depend only on $\hat\omega=\omega/e$, so a uniform scaling $\omega\mapsto s\omega$ leaves $\hat\omega$ unchanged. The sums $F_{xx}$ and $F_{x\mu}$, hence $\tilde{\mathcal{E}}_h=\log F_{xx}-2\log F_{x\mu}$ and the softmax weights $\alpha,\beta$, are unchanged. By the chain rule, $\partial_{x_t}\tilde{\mathcal{E}}_h = (\partial \hat g/\partial x_t)^\top \partial_{\hat\omega_t}\tilde{\mathcal{E}}_h$, and $\partial \hat g/\partial x_t=(1/e)\,\partial g/\partial x_t$ cancels the physical scaling inside $\partial_{\hat\omega_t}$, yielding identical gradients (cf. Lemma~\ref{lem:bounds}).
\end{proof}

With this last proof, we formulate the final ergodic trajectory optimization as
    \begin{equation} \label{eq:log_logdt_erg_opt}
        \min_{x, \log(\Delta)} \tilde{\mathcal{E}}_h(x ; \mu) \quad\text{s.t.  } \forall t \, f(x_{t+1},x_t, \Delta_t) = 0, h(x) \le 0.
    \end{equation}
    where $h$ is annealed given~\eqref{eq:annealing} per iteration. 
    In implementation, we solve the above optimization using an Augmented Lagrangian method~\cite{birgin2014practical, rockafellar1973}.


\section{RESULTS} \label{sec:results}

\subsection{Environmental Scale in Ergodic Kernel-Based Methods}


    We first evaluate the dependence of existing kernel-based methods on the length parameter by conducting an ablation study (shown in Fig.\ref{fig:kernel_comp}) on the variation of length parameter selection for the traditional ergodic MMD solver and the fast ergodic kernel solver provided in \cite{sun2024}. 
    Here, we find that variations in the kernel length parameter can cause drastic changes in how the optimization criteria are defined, such as the visitation footprint in MMD. 
    Given that the kernel length parameter's primary function is to denote the distance between points, different scales will require different selections of $h$ for proper representation of the spatial data, but modification of the length scale to adequately represent the spatial scale of the environment comes at the cost of altering the ergodic trajectory's behavior as well. 
    These results show that existing methods' reliance on spatial scale leads to a tedious trial-and-error process for appropriate hyperparameter selection for each new environment explored, which nullifies the geometrical generalization that these methods offer.  

    \begin{figure}[h]
    \centering
    \vspace{5pt}
    \includegraphics[width=\linewidth]{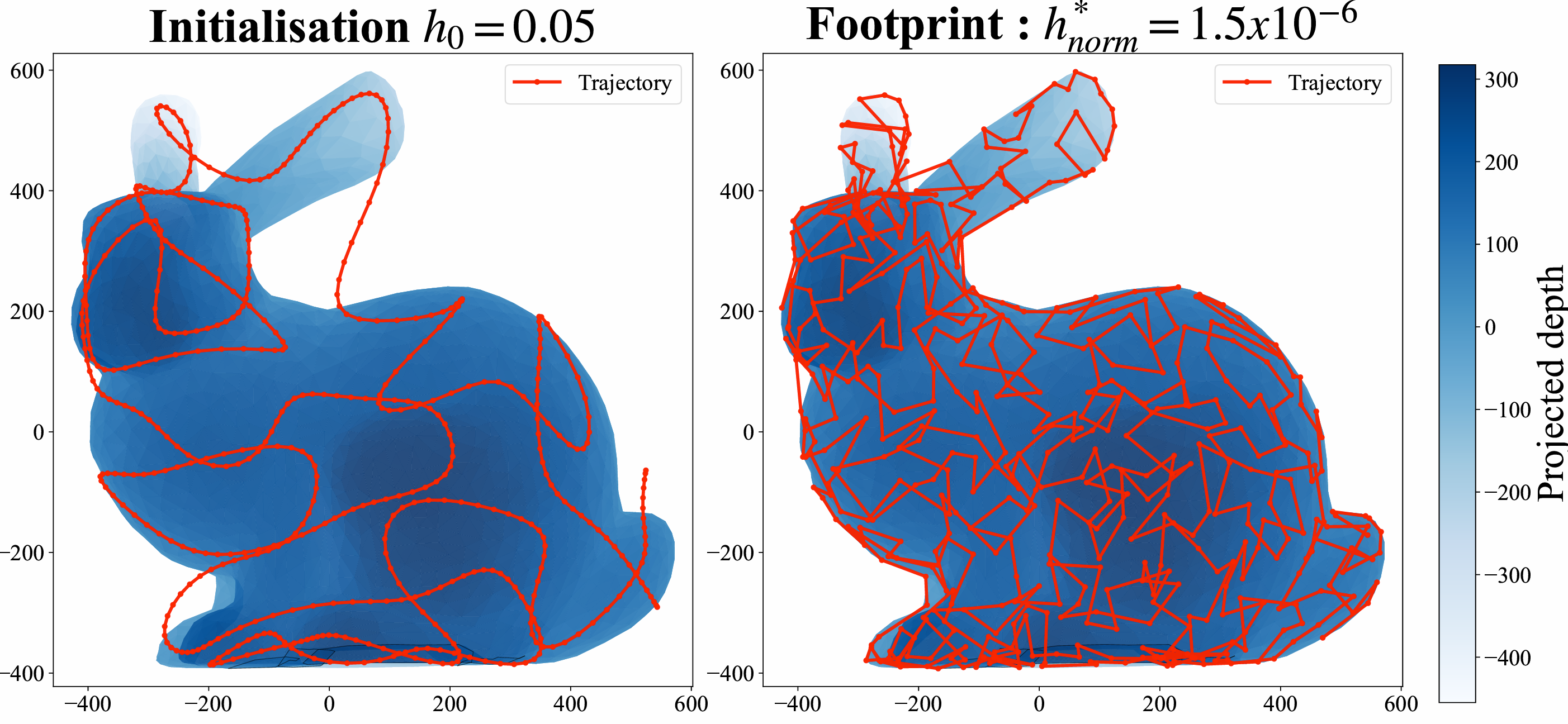}
    \vspace{-15pt}
    \caption{\textbf{Bandwidth annealing preserves a physical footprint and stabilizes optimization at scale.} Left: starting with a large normalized bandwidth $h_{0}=0.05$ yields broad similarity and well-conditioned gradients on a large domain. Right: after continuation to the \emph{target physical} footprint of $1\,\mathrm{m}$ (which corresponds to $h_{\mathrm{norm}}^{\star}=1.5\times10^{-6}=h_{\mathrm{phys}}^{\star}/e^{2}$), the optimized trajectory (red, $T=500$ knots) achieves fine, proportional coverage of the bunny while constraints are enforced in physical units; domain extent $e=1000\,\mathrm{m}$.}

    \label{fig:annealing_comparison}
    \end{figure}
    
\begin{figure}[t]
    \centering
    \includegraphics[width=1.0\linewidth]{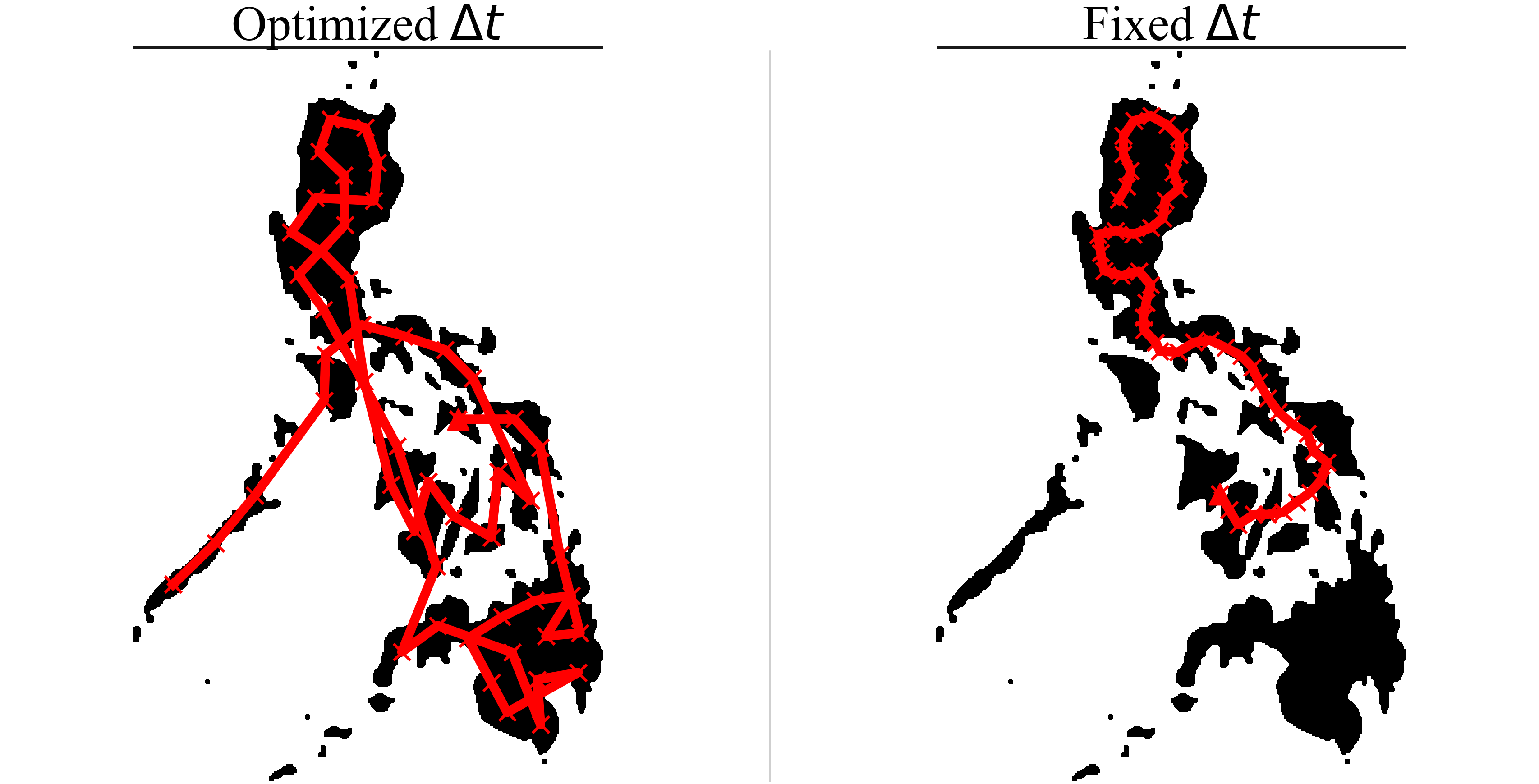}
    \vspace{-6pt}
    \caption{\textbf{Adaptive vs.\ fixed time steps on a sparse, multi-scale target.}
    Philippines silhouette (black) with identical trajectory length, \(T=50\). 
    \emph{Left} (Optimized \(\Delta t\)): time is concentrated over dense archipelagos; the trajectory refines coverage on many small islands and traverses open water quickly.
    \emph{Right} (Fixed \(\Delta t\)): uniform spacing oversamples empty water and undersamples intricate regions, yielding poorer effective resolution on the islands.}
    \label{fig:compare_dt}
    \vspace{-10pt}
\end{figure}

    Likewise, we show that our method is able to adapt 1) its footprint on a large-scale search problem in Fig.~\ref{fig:annealing_comparison} while preserving a more restrictive sensor footprint determined by the length-scale $h$. 
    In addition, having $\Delta_t$ be a free variable allows us to plan coverage trajectories over highly sparse domains like a collection of islands in Fig.~\ref{fig:compare_dt}. Here, we see that with a fixed $\Delta_t$, trajectory placement can become challenging over the spatially sparse map which an adaptive $\Delta_t$ allows differential constraints to adapt while satisfying coverage performance. 
    
    
\subsection{Quantitative Analysis and Spatial Scalability}

\begin{figure}[t]
    \centering
    \vspace{5pt}
    \includegraphics[width=\linewidth]{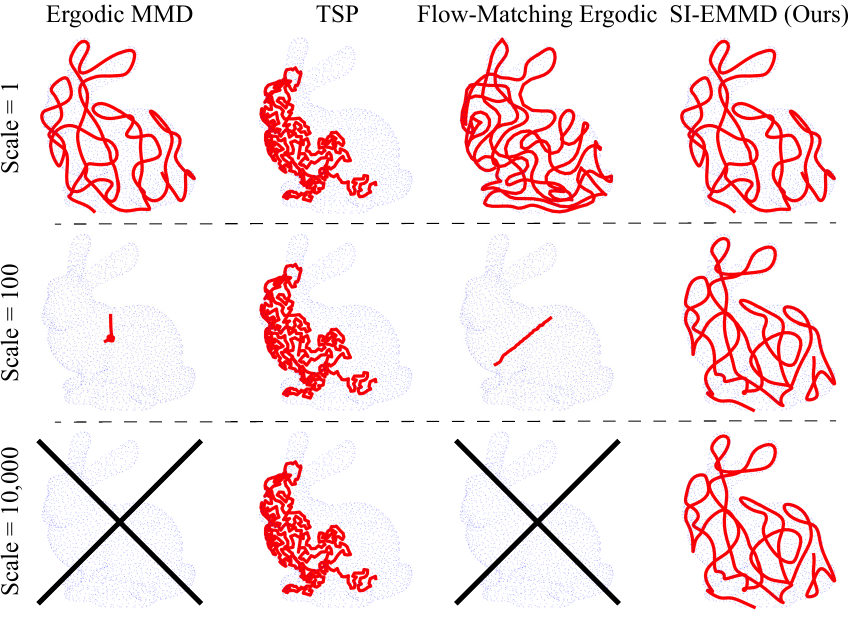}
    \vspace{-15pt}
    \caption{\textbf{Comparison of Scale-Invariant EMMD to State-of-the-Art Methods.} Quantitative comparison of our scale-agnostic method against state-of-the-art coverage planners on the Stanford bunny mesh (2,503 pts). Trajectories are constrained to an identical maximal length that scales proportionally with environment size. Where existing ergodic exploration methods fail to adapt to vastly different spatial scales, and TSP solvers struggle to generate trajectories that prioritize coverage in length-constrained settings, our method consistently achieves high coverage and maintains numerical stability across scales, while maintaining similar optimization times and coverage quality to ergodic MMD trajectories at scale = 1.}
    \label{fig:compare_3d}
    \vspace{-15pt}
    \end{figure}

    Next, we compare the performance of our scale-agnostic ergodic trajectory optimization approach against several state-of-the-art coverage planners over a 3D bunny mesh, as shown in Fig.\ref{fig:compare_3d}. 
    Coverage trajectories were generated using the traditional ergodic MMD, traveling salesperson, flow-matching ergodic \cite{sun2025}, and scale-agnostic EMMD planners across three different spatial scales (1, 100, and 10,000), each with identical maximum trajectory length and initial position constraints for the given scale. 

    Our scale-agnostic formulation consistently achieved high coverage and generates smooth trajectories across all scales. In addition, the performance of our approach at each scale was nearly identical to that of the canonical ergodic MMD solver at scale = 1, which verifies the scale-invariance property. In contrast, as the spatial scale increased, the canonical solver's performance degraded significantly at scale = 100 and failed to converge at scale = 10,000. 

    The flow-matching ergodic solver achieved high coverage within the normalized domain, but similar to the traditional MMD approach, its performance degraded with increasing scale. This is because the flow-matching ergodic solver is based on Sinkhorn divergence, which characteristically interpolates between MMD and optimal transport metrics \cite{sun2025}. 
    Similar to the traditional MMD metric, the Sinkhorn divergence metric naturally struggles with optimization at larger spatial scales. 
    In addition, the Sinkhorn divergence at vastly different spatial scales incurs a computational cost that scales linearly with the scale of the search space. 
    In particular, the Sinkhorn divergence relies more on its leverage of the Wasserstein distance, which assists with coverage performance at small variations in scale at the cost of slower computational speeds. 

    The numerical instability of Sinkhorn divergence at vastly larger scales, such as 10,000, yielded few tractable gradients with which to follow, which decreased computational requirements at the cost of a viable coverage trajectory. 
    However, at scales where the Sinkhorn divergence metric did not fail, its computational variance to spatial scale and the number of computed samples resulted in infeasibly-long computation times which we show in Table~\ref{tab:quant_analysis}.

\begin{table}[h]
\centering
\vspace{5pt}
\caption{Quantitative Analysis: Comparison Against Existing Methods in 3D}
\label{tab:quant_analysis}
\setlength{\tabcolsep}{3pt} 
\begin{tabular}{lccc}
\toprule
\textbf{Metric} & \textbf{Scale 1} & \textbf{Scale 100} & \textbf{Scale 10,000} \\
\midrule
\multicolumn{4}{l}{\textbf{EMMD}} \\
\quad Coverage (\%) & 91.69 & 0.52 & 0.00 \\
\quad Comp. Time (sec) & 39.66 $\pm$ 0.32 & 39.75 $\pm$ 0.24 & 31.84 $\pm$ 0.24 \\
\midrule
\multicolumn{4}{l}{\textbf{TSP}} \\
\quad Coverage (\%) & 42.31 & 42.39 & 42.39 \\
\quad Comp. Time (sec) & 0.015 $\pm$ 0.003 & 0.014 $\pm$ 0.001 & 0.013 $\pm$ 0.001 \\
\midrule
\multicolumn{4}{l}{\textbf{Flow-Matching}} \\
\quad Coverage (\%) & 98.60 & 0.88 & 0.00 \\
\quad Comp. Time (sec) & 1249.45 $\pm$ 228.5 & 60.60 $\pm$ 6.57 & 28.95 $\pm$ 0.30 \\
\midrule
\multicolumn{4}{l}{\textbf{SI-EMMD (Ours)}} \\
\quad Coverage (\%) & 91.69 & 91.09 & 91.09 \\
\quad Comp. Time (sec) & 40.49 $\pm$ 0.31 & 40.43 $\pm$ 0.25 & 39.47 $\pm$ 0.41 \\
\bottomrule
\end{tabular}
\end{table}

    Despite consistent performance across varying scales, the traveling salesperson solver demonstrated suboptimal coverage over the mesh due to its inability to globally select which point to visit. 
    Although these solvers can guarantee full coverage of all reachable space given a sufficiently long trajectory, they do not intrinsically prioritize broad visitation within a constrained trajectory length. 

\begin{figure}[h]
    \centering
    \includegraphics[width=\linewidth]{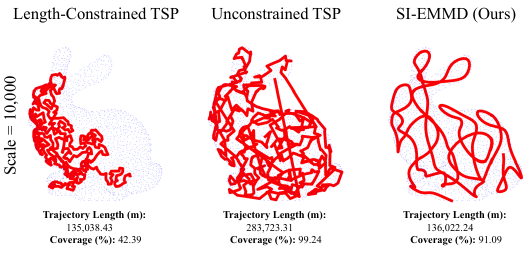}
    \vspace{-15pt}
    \caption{\textbf{Constraint Violations for Similar Coverage with TSP.} Trajectory optimization of three different methods on the Stanford bunny model. The standard Traveling Salesperson solver (TSP) with an imposed trajectory distance constraint achieves only 42.39\% coverage due to TSP's general incompatibility with imposed constraints. When the length constraint is removed and the bunny mesh is uniformly sampled to the number of control knots (trajectory points), TSP achieves 99.24\% coverage at the cost of more than double the trajectory length. The proposed scale-invariant metric achieves 91.09\% coverage at significantly less trajectory length while accommodating physical constraints.}
    \label{fig:tsp_fair}
    \vspace{-10pt}
\end{figure}

    For a comparative analysis of unconstrained coverage, we compare TSP's unconstrained coverage to our proposed method by uniformly sampling the bunny mesh to the desired length of trajectory control knots (500). 
    As shown in Fig.\ref{fig:tsp_fair}, the unconstrained TSP path achieved near-perfect coverage, but at the expense of a trajectory more than twice as long as any length-constrained version. 
    In addition, this trajectory proved dynamically infeasible due to its jagged and abrupt movements between adjacent points.
    In contrast, our scale-invariant EMMD solver achieves comparable coverage while strictly adhering to smoothness and velocity constraints, thus simultaneously ensuring trajectory feasibility and preserving the proportional coverage guarantees of ergodic methods. 

\subsection{Real-World Validation and Applications}

\begin{figure}[h]
    \centering
    \includegraphics[width=\linewidth]{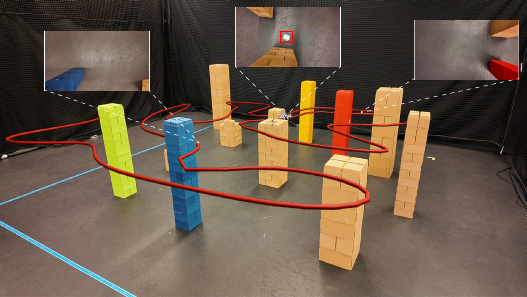}
    \vspace{-15pt}
    \caption{\textbf{Real-World Autonomous Object Search Using Scale-Agnostic Optimization.} Our proposed optimizer is validated on a physical Crazyflie drone to perform a real-world coverage task on a millimeter-scale 'block city'. Snapshots from the drone's downward-facing view confirm the successful execution of the coverage task and discovery of a stationary target that is hidden between buildings.}
    \label{fig:drone_xplor}
\end{figure}

    In order to validate the practical utility of our method, we deployed the scale-invariant ergodic MMD optimizer on a real-world robotic platform. 
    A Crazyflie 2.1 drone is placed in a millimeter-scale 'block city' within a 3m x 3m x 1 m flight cage (as shown in Fig.\ref{fig:drone_xplor}) and tasked with uniformly locating a stationary sphero hidden between the buildings. 
    Despite the significant spatial scale (9,000,000,000 $mm^3$ search volume), our optimization method generated a effective coverage trajectory while respecting the drone's strict dynamical constraints. 
    The generated ergodic trajectory successfully guided the drone in between the buildings according to a set of control barrier functions placed about the buildings and allowed the drone to successfully locate the target. 
    This experiment confirms the feasibility of our coverage planner to integrate with physical systems and execute complex exploration tasks with high precision. 

    To further display the physical applications of our method to large-scale industrial tasks, we generated a coverage trajectory over the hull of a naval ship with a length of 190 m (see Fig.\ref{fig:abstract}). 
    At ports, trajectories of this type can be tracked by underwater robots in order to clean the underside of ships without the need for costly and time-consuming approaches to ship maintenance, such as lifting the boats out of the water and manually washing and scrubbing the hull. 
    The enablement of autonomous systems to perform basic coverage tasks over objects of arbitrary geometric structure and physical size yields the potential to create substantial economic and time savings in fields of maritime maintenance, materials processing, search-and-rescue, and even global-scale monitoring tasks. 

\section{CONCLUSION} \label{sec:conclusion}

    We introduce a scale-invariant ergodic MMD trajectory optimizer that alleviates numerical challenges of kernel-based ergodic methods to spatial scale. 
    Our approach leverages a novel MMD formulation that guarantees ergodic coverage over the reachable search space within a normalized domain while preserving physical constraints at their true scale. 
    Furthermore, our method's performance is evaluated in a variety of physical implementations and simulations which showcase its light computational requirements, coverage quality, and resilience across diverse environmental scales. 

\bibliographystyle{IEEEtran}
\bibliography{references}

\end{document}